\documentclass[letterpaper, 10 pt, conference]{ieeeconf}  

\IEEEoverridecommandlockouts                              

\usepackage{graphics} 
\usepackage{epsfig} 
\usepackage{mathptmx} 
\usepackage{times} 
\usepackage{amsmath} 
\usepackage{amssymb}  
\usepackage{siunitx}
\usepackage{booktabs}
\usepackage{multirow}
\usepackage{subcaption}
\usepackage[hidelinks]{hyperref}

\title{\LARGE \bf
Dynamic Shape Control of Soft Trunk Robots}

 \author{Hehui Zheng$^{1,\dagger}$, Marc Stampanoni$^{1,\dagger}$, Roshan S. Kaundinya$^{2,\dagger}$,\\  Ronan Hinchet$^{1}$, George Haller$^{2}$, and Robert K. Katzschmann$^{1}$%
\thanks{$\dagger$ These authors contribute equally}
 \thanks{$^{1}$Hehui Zheng, Marc Stampanoni, Ronan Hinchet, and Robert K. Katzschmann are with Soft Robotics Lab, ETH Zurich, Zurich, Switzerland
         {\tt\small \{\href{mailto:zhengh@ethz.ch}{zhengh}, \href{mailto:mstampanoni@ethz.ch}{mstampanoni}, \href{mailto:rhinchet@ethz.ch}{rhinchet}, \href{mailto:rkk@ethz.ch}{rkk}\}@ethz.ch}}%
 \thanks{$^{2}$Roshan S. Kaundinya and George Haller are with the Institute for Mechanical Systems, ETH Zurich, Zurich, Switzerland
         {\tt\small \{\href{mailto:sroshan@ethz.ch}{sroshan}, \href{mailto:georgehaller@ethz.ch}{georgehaller}\}@ethz.ch}}%
 }

\begin{document}


\addtocounter{figure}{0}
\makeatletter
\let\@oldmaketitle\@maketitle%
\renewcommand{\@maketitle}{
  \@oldmaketitle%
  \begin{center}
  \vspace{10pt}
    \includegraphics[width=177.8mm]{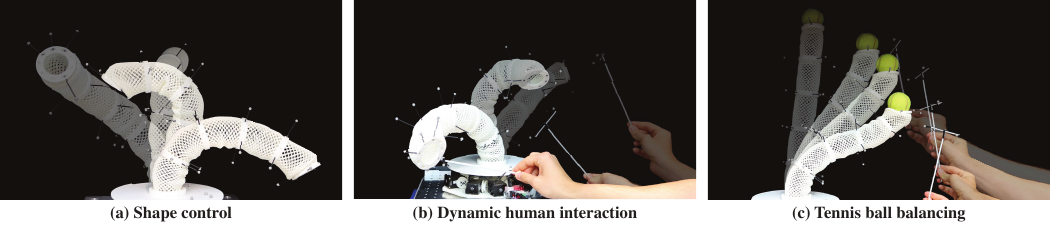}
    \captionsetup{font=footnotesize}
    \captionof{figure}{Closed-loop full-body shape control of a five-segment trimmed-helicoid soft trunk.
    An MPC controller built on an adiabatic spectral submanifold (aSSM) model learned from hardware data drives the trunk on three tasks: \textbf{(a)} tracking a target shape, \textbf{(b)} reshaping under human interaction with a hand-held wand, and \textbf{(c)} following the wand while holding the tip level to balance a tennis ball.}
    \label{fig:hero}
    \vspace{-2.0\baselineskip}
  \end{center}\bigskip
}
\makeatother
\maketitle
\setcounter{figure}{1}

\begin{abstract}
Controlling the shape profile of soft trunk robots extends their reach into complex regions of the workspace and enhances their use in practical applications. Recent advances in soft robot design have enabled soft trunk robots to achieve greater flexibility in conforming to complex shapes. Specifically, trunk robots whose geometries are designed as helicoids have been shown to enable extreme shape-changing capabilities. Their compliance makes them highly nonlinear, which in turn makes shape control difficult. On similar tendon-driven trunk robots, model-based controllers based on the theory of adiabatic spectral submanifolds (aSSMs) have achieved high accuracy in closed-loop end-effector tracking. We extend this approach by learning an aSSM model of the full body of a trimmed-helicoid soft trunk robot from data and use this in a model predictive control (MPC) scheme to demonstrate a range of shape control tasks on hardware. On closed-loop dynamic shape regulation, our method outperforms the available linear baselines by a factor of 10. In human-in-the-loop interactive demos, our method completes section-by-section shaping and balances a tennis ball for over \SI{25}{\second}, while the linear baselines fall short. A model-free reinforcement learning policy trained for \SI{12}{\hour} on hardware fails on the same balancing task. These results establish aSSM-based MPC as a practical route to full-body shape control on real soft hardware.

\end{abstract}

\section{Introduction}

\begin{figure*}[t]
  \centering
  \includegraphics[width=\textwidth]{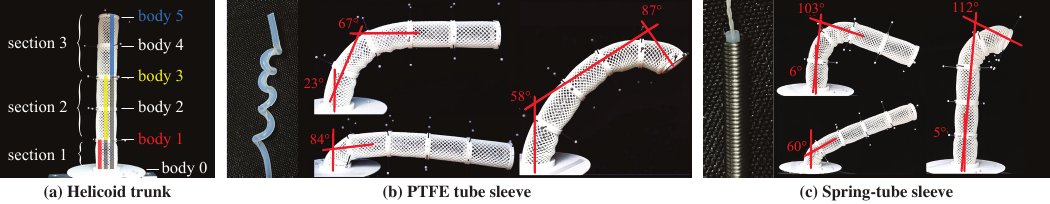}
  \caption{
  Hardware platform and the effect of tendon routing on inter-section coupling. 
  \textbf{(a)} Five trimmed-helicoid segments stacked in series, with marker body~0 at the base and bodies~1--5 at the distal end of each segment.
  Colored strips illustrate the three tendon groups driving the sections, grey marks the sleeved portion routed through the segments below.
  \textbf{(b)} With PTFE tubes, the sleeves plastically deform (left), giving up to $58^\circ$ of uncommanded deflection.
  \textbf{(c)} Nesting each tube in an extension spring (left) reduces the largest uncommanded deflection to $6^\circ$ while increasing the commanded bend of section~3 from $87^\circ$ to $112^\circ$.
  Section~2 uses a $0.8$/$1.2$\,mm tube in a $2.0$\,mm spring, section~3 a $0.6$/$1.6$\,mm tube in a $3.0$\,mm spring.
  }
  \label{fig:hardware}
  \vspace{-1em}
\end{figure*}

Soft trunk robots are designed to navigate sensitive environments, where their entire bodies must deform to operate successfully \cite{rus2015design, laschi2016soft}. Examples include maneuvering in unstructured environments, such as during endoscopy surgeries \cite{Burgner-Kahrs_medical}, fuel tank inspections \cite{rosalind_2207}, and disaster site inspections \cite{tsukagoshi2001active}. For such applications, a controller needs to track the trunk's shape accurately. Close to the undeformed state, the trunk's shape dynamics are linear, but as the trunk deviates from this state, the dynamics tend to become highly nonlinear.

Modeling the nonlinear dynamics of the trunk's shape is essential to enable shape control via model predictive control (MPC) schemes. Working directly on hardware poses additional challenges, such as no access to the governing equations and only limited access to observable state-space coordinates through data. In such cases, data-driven modeling via invariant manifolds (IMs) that capture the dominant shape dynamics is a robust strategy. Moreover, the dominant dynamics of soft robots are expected to lie on a low-dimensional IM due to an inherent time-scale separation: their intrinsic vibrations decay much faster than their operating speeds.

Specifically, spectral submanifolds (SSMs) and adiabatic SSMs (aSSMs) are the IMs that contain these dominant dynamics. The former models soft robot dynamics locally in the trunk's workspace \cite{alora2025discovering}, whereas the latter models the workspace globally \cite{Kaundinya2026}. Both methods have previously been used only to track the end-effector of the trunk, and the aSSM modeling approach has used additional information to avoid obstacles across the full workspace \cite{kaundinya2026precision}. Recently, \cite{karakai2026orientationcontrolsoftrobots} has shown that aSSMs can be used to accurately control the end effector and its orientation on a soft trunk simulator. Motivated by these successes, we learn a low-dimensional aSSM-reduced model for a helicoid soft trunk robot and use it in an MPC scheme to demonstrate full-body shape control.


In this work, we contribute by:

\begin{itemize}

    \item learning an aSSM-reduced model of the full body from hardware data and deploying it in an MPC scheme, extending an approach previously applied only to end-effector tracking;
    
    \item demonstrating shape control on a range of tasks: dynamic shape tracking, section-by-section shaping and balancing a tennis ball in human-in-the-loop demos. The aSSM-based MPC outperforms the linear baselines by an order of magnitude, and a model-free reinforcement learning (RL) policy trained for 12 hours on hardware fails to complete the balancing task;

    \item redesigning the tendon sleeves of the helicoid trunk platform, which reduces the largest inter-section coupling from \SI{58}{\degree} to \SI{6}{\degree} of uncommanded deflection and makes the sections independently commandable.

\end{itemize}

\section{Related Work}

Traditional modeling methods use knowledge of material laws and, under kinematic assumptions, develop simple modeling strategies to describe the shape dynamics of soft robots. 
For soft trunks, piecewise constant curvature (PCC) \cite{DellaSantina2023_mag, Katzschmann_cc} assumes that the trunk's body can be split into segments of fixed length bending under constant curvature, and piecewise constant strain (PCS) \cite{renda18} relaxes the fixed-length assumption by taking the deformation strains across the segment as constant. 
On hardware, neither is readily implementable, since the material laws and their kinematics are unknown. 
Where the design parameters and model geometry are known, \cite{guan2023trimmed} approximates a soft trunk as a finite element model (FEM) and derives a PCC-based model from it for inverse-kinematics control.
However, obtaining such an FEM is challenging for soft robotic platforms with distinct geometries, actuation, and material properties. 

A general solution is to learn dynamic models from data. To that end, linear methods, assuming the dynamics follow linear evolution, fit the best linear system to data collected on hardware. 
With a specific focus on shape control, \cite{Adibnazari2025} compares Lagrangian Operator Inference (LOpInf) \cite{sharma_LpOpInf}, dynamic mode decomposition with control (DMDc) \cite{Proctor2016}, and Eigensystem realization algorithm (ERA) \cite{juang85}. 
For the soft robotic fish considered in their work, LOpInf performed best in simulation, but on hardware all three methods struggled to track a large reference due to the inherent nonlinearity. 

That nonlinearity is unavoidable on a helicoid trunk, and recent work on this platform addresses it in three ways. 
\cite{zhang2026printedhelicoidsembeddedair} gives design rules for 3D printing a trimmed helicoid with a model of the printed structure, validated only in open loop. 
\cite{wong2026closedformclfcbfcontrollerwholebody} derives closed-form equations of motion for a helicoid trunk and exploits them for whole-body obstacle avoidance, a shape-level objective, though one that presumes an analytical model rather than one identified from hardware.
\cite{fang_2026}, instead, learns a neural-network dynamics model for model-based RL, recovering the nonlinearity at the cost of a high-dimensional model, and tracks 2D targets over paths short relative to the trunk length.

As an alternative to fitting nonlinear models from data, slow spectral submanifolds (SSMs) exploit the invariant structure the dynamics already possess. 
\cite{alora2025discovering} extract them locally near an equilibrium to control a hardware Diamond robot. 
The restriction to a neighborhood of a single equilibrium is lifted in \cite{Kaundinya2026}, using the time-scale separation common to soft robotic control to extract the dominant dynamics globally onto adiabatic spectral submanifolds (aSSMs) through an extension of the local theory \cite{haller24_wa}. 
Across a range of soft robotic simulators, aSSMs identified from data support accurate trajectory tracking within an MPC scheme.
On hardware, aSSM-reduced models enable precise 3D target tracking on helicoid robots over paths comparable to its body length \cite{kaundinya2026precision} and orientation control has been addressed in simulation \cite{karakai2026orientationcontrolsoftrobots}. 
Shape control, in which the configuration of the entire body rather than a single point or frame is regulated, has not yet been considered. We address this problem in the next sections.

\section{Methodology}

\subsection{Robot platform}
\label{sec:platform}
The robot is a soft trunk of five trimmed-helicoid segments stacked in series (Fig.~\ref{fig:hardware}a), each following the geometry of~\cite{guan2023trimmed}, with a helical pitch of \SI{45}{\degree}, 20 helicoids, an outer radius of \SI{32}{\milli\metre}, a wall thickness of \SI{5}{\milli\metre}, a single-helicoid thickness of \SI{1.4}{\milli\metre}, and a TPU material hardness of 95A. 
This geometry yields a large set of reachable shapes from a short stack, and the compliance  that enables this also makes the actuation-to-shape map strongly nonlinear. 
The trunk is \SI{500}{\milli\meter} long unloaded and \SI{170}{\gram} in mass.

\begin{table}[t]
\centering
\caption{Actuation limits and inter-section coupling for the two tendon routing designs. Each section is driven alone to its tendon limit, and the coupled deflection is the uncommanded bend induced below the actuated section. Limits are set conservatively per design to avoid mechanical failure.}
\label{tab:routing}
\small
\setlength{\tabcolsep}{4pt}
\begin{tabular}{llcccccc}
\toprule
 &  & \multicolumn{2}{c}{Limit (ticks)}
    & \multicolumn{3}{c}{Deflection} \\
\cmidrule(lr){3-4}\cmidrule(lr){5-7}
Routing & Sec. & Tendon & Sum & Act. & Coup. & Total \\
\midrule
\multirow{3}{*}{PTFE tube}
 & 1 & 3800 &  5000 &  $84^\circ$ &  $0^\circ$ &  $84^\circ$ \\
 & 2 & 5000 &  7000 &  $67^\circ$ & $23^\circ$ &  $90^\circ$ \\
 & 3 & 8000 & 10\,000 & $87^\circ$ & $58^\circ$ & $145^\circ$ \\
\midrule
\multirow{3}{*}{Spring--tube}
 & 1 & 2500 &  3500 &  $60^\circ$ &  $0^\circ$ &  $60^\circ$ \\
 & 2 & 5000 &  7000 & $103^\circ$ &  $6^\circ$ & $109^\circ$ \\
 & 3 & 7000 & 11\,200 & $112^\circ$ & $5^\circ$ & $117^\circ$ \\
\bottomrule
\end{tabular}
\end{table}

\subsubsection{Actuation}
Nine tendons drive the trunk in three independently actuated sections of three tendons each, spaced $120^\circ$ about the axis, so every section is commanded in two bending directions and one axial contraction. 
Section~$k$ is actuated by the motor group $G_k$, with $G_1=\{M_1,M_4,M_7\}$, $G_2=\{M_2,M_5,M_8\}$ and $G_3=\{M_3,M_6,M_9\}$.
Section~1 spans segment 1, section~2 segments 2--3 and section~3 segments 4--5. 
Each tendon is driven by a Dynamixel XM430-W350-R servo in extended position mode through a \SI{19.5}{\milli\metre}-diameter spool.
The robot is therefore commanded by nine motor angles $\mathbf{u}\in\mathbb{R}^{9}$, expressed in encoder ticks. 
Per-tendon and per-section limits (Table~\ref{tab:routing}) are set conservatively to prevent mechanical failure or sustained large deflections. 

\subsubsection{Sensing}
Six rigid marker bodies are tracked by an eight-camera Qualisys motion capture system at \SI{100}{\hertz}: body~0 on the fixed base mount, and bodies $1,\dots,5$ at the distal end of each segment. 
Poses are expressed relative to the base body~0, and the positions $\mathbf{p}_i\in\mathbb{R}^{3}$ and orientation quaternions $\mathbf{q}_i\in\mathbb{S}^{3}\subset\mathbb{R}^{4}$ of the five moving bodies are concatenated into the pose vector
\begin{equation}
  \mathbf{s} = \big[\mathbf{p}_1^{\!\top},\mathbf{q}_1^{\!\top},
                    \dots,
                    \mathbf{p}_5^{\!\top},\mathbf{q}_5^{\!\top}
              \big]^{\!\top} \in \mathbb{R}^{35}.
  \label{eq:pose}
\end{equation}

\subsubsection{Tendon routing and inter-section decoupling}
The tendons actuating sections~2 and~3 are routed through the segments below them, so without constraints, actuating an upper section would also bend the helicoid segments beneath it.
Independent routing through Bowden cables decouples the sections and enlarges the reachable workspace~\cite{guan2023trimmed}, and we initially adopted the same approach with a polytetrafluoroethylene (PTFE) tube fixed to the segments each tendon passes through.
However, PTFE tubes deformed plastically after several hours of operation, and the resulting friction exceeded the restoring force of the structure, leaving the trunk bent. We therefore insert each PTFE tube into a closely wound extension spring, which preserves the low tendon-to-sleeve friction while resisting axial shortening. 
Driving each section alone to its tendon limit (Fig.~\ref{fig:hardware}b--c; Table~\ref{tab:routing}), the largest uncommanded deflection falls from $58^\circ$ to $6^\circ$. 

\subsection{Adiabatic spectral submanifolds for MPC}\label{sec:assm_theory}
\label{subsec:theory_MPC}

The equations of motion of soft robots take the abstract form,
\begin{equation}
\label{eq:eom_MPC}
\dot{\mathbf{x}} = \mathbf{F}(\mathbf{x}, \mathbf{u}(t)), \qquad \mathbf{x} \in \mathcal{X} \subseteq \mathbb{R}^{n}, \quad \mathbf{u} \in \mathcal{U} \subseteq \mathbb{R}^{n_u},
\end{equation}
with $\mathbf{x}$ the phase space of the robot, $\mathbf{u}$ the control inputs and $\mathcal{U}$ the admissible control input set. 

When the control input varies slowly with a small, bounded fast deviation,
\begin{equation}
\label{eq:control_action_split}
\mathbf{u}(t) = \mathbf{u}^s(\epsilon t) + \mathbf{u}^d(t), \qquad \|\mathbf{u}^d(t)\| < \Delta, \quad 0 < \epsilon \ll 1,
\end{equation}
\cite{haller24_wa} and \cite{Kaundinya2026} prove that system (\ref{eq:eom_MPC}) admits a time-varying, normally attracting slow manifold $\mathcal{L}_\epsilon$ and an adiabatic spectral submanifold (aSSM) $\mathcal{A}_\epsilon$ anchored to it. This holds if a spectral gap separates the $d$ slowest eigenvalues of the linearized problem in the $\epsilon =0$ limit of (\ref{eq:eom_MPC}) from the rest. The critical limit $\mathcal{L}_0$ of $\mathcal{L}_\epsilon$ as $\epsilon \to 0$ is used for aSSM $\mathcal{A}_\epsilon$ construction from data \cite{Kaundinya2026,kaundinya2026precision}.

Hardware provides only a sparse positional observable $\mathbf{s} \in \mathcal{S} \subseteq \mathbb{R}^{n_s}$. We therefore embed the $d$-dimensional aSSM with the delay embedding $\mathbf{y}(t) = \boldsymbol{\mu}(\mathbf{s}(t) )= \big(\mathbf{s}(t), \mathbf{s}(t-\tau), \dots, \mathbf{s}(t-p\tau)\big) \in \mathbb{R}^{n_s(p+1)}$, where $\tau > 0$ and $p \geq \lfloor (2d+1)/n_s \rfloor$ by Takens' theorem \cite{takens81}.

We learn the leading-order aSSM-reduced model
\begin{align}
\label{eq:aSSM_leading_MPC}
\mathbf{r} &= \mathbf{V}^{\mathrm T}(\mathbf{u}(t))\, \boldsymbol{\mu}\big(\mathbf{s} - \bar{\mathbf{s}}(\mathbf{u}(t))\big), \nonumber \\
\dot{\mathbf{r}} &= \mathbf{R}\big(\mathbf{r}, \bar{\mathbf{s}}(\mathbf{u}(t))\big), \\
\mathbf{s} &= \mathbf{W}\big(\mathbf{r}, \bar{\mathbf{s}}(\mathbf{u}(t))\big) + \bar{\mathbf{s}}(\mathbf{u}(t)), \nonumber
\end{align}
where $\bar{\mathbf{s}}(\mathbf{u})$ is the critical manifold in observable coordinates, $\mathbf{V}(\mathbf{u}(t))$ spans the time-varying slow subspace, $\mathbf{r}$ is the reduced coordinate, $\mathbf{W}(\cdot)$ is the aSSM parametrization, and $\mathbf{R}(\cdot)$ is the reduced dynamics.

MPC needs a discrete-time prediction model over its horizon. Integrating \eqref{eq:aSSM_leading_MPC} over a step $\Delta t$ under a zero-order hold on $\mathbf{u}_k$ gives
\begin{align}
\label{eq:aSSM_discrete_MPC}
\mathbf{r}_k &= \mathbf{V}^{\mathrm T}(\mathbf{u}_k)\, \boldsymbol{\mu}\big(\mathbf{s}_k - \bar{\mathbf{s}}(\mathbf{u}_k)\big), \nonumber \\
\mathbf{r}_{k+1} &= \mathbf{R}_d\big(\mathbf{R};\, \mathbf{r}_k, \bar{\mathbf{s}}(\mathbf{u}_k), \Delta t\big), \\
\mathbf{s}_{k+1} &= \mathbf{W}\big(\mathbf{r}_{k+1}, \bar{\mathbf{s}}(\mathbf{u}_k)\big) + \bar{\mathbf{s}}(\mathbf{u}_k), \nonumber
\end{align}
with $\mathbf{R}_d(\cdot)$ the discrete map approximated numerically from $\mathbf{R}(\cdot)$, so the same identified model can be used at different step sizes without refitting. This is the model we propagate over the prediction horizon inside the MPC optimization.

\subsection{Data-driven identification of aSSMs}\label{subsec:datadriven_MPC}

We collect the robot's response to step control inputs in the observable space. The step inputs are generated by jumping from one control input to the next across a set of $N = 20,000$ static inputs, each held for $t_{\mathrm{decay}}$ seconds, giving $1.79\times10^{6}$ data points sampled at $100$~Hz. We sample the static input set with a Sobol sequence \cite{sobol_sequence}. Compared to uniform random sampling, the Sobol sampling method requires fewer samples to cover the reachable workspace. Passing these control inputs gives static pairs $\{\mathbf{s}^i_{\mathrm{static}}, \mathbf{u}^i_{\mathrm{static}}\}_{i=1}^{N}$ and their accompanying decay trajectories $\{\mathbf{s}^i_{\mathrm{decay}}, \mathbf{u}^i_{\mathrm{decay}}\}_{i=1}^{N}$.

\subsubsection{Learning the critical manifold}

We learn $\bar{\mathbf{s}}(\mathbf{u})$ as a nonlinear static map by regressing the static configurations onto their control inputs,
\begin{equation}
\label{eq:learn_cm_MPC}
\bar{\mathbf{s}}^{\star} = \operatorname*{arg\,min}_{\bar{\mathbf{s}}}\ \sum_{i=1}^{N} \big\| \mathbf{s}^i_{\mathrm{static}} - \bar{\mathbf{s}}(\mathbf{u}^i_{\mathrm{static}}) \big\|^2.
\end{equation}
We seek the critical manifold as a radial-basis-function expansion over $M$ centers $\{\mathbf{u}_j\}_{j=1}^{M}$ with the linear kernel,
\begin{equation}
\label{eq:cm_rbf_MPC}
\bar{\mathbf{s}}(\mathbf{u}) = \sum_{j=1}^{M} \mathbf{w}_j\, \|\mathbf{u} - \mathbf{u}_j\| + \mathbf{c}_0,
\end{equation}
where $\mathbf{w}_j$ are the kernel weights and $\mathbf{c}_0$ is a constant offset. The expansion is linear in its coefficients, so the regression reduces to a linear least-squares problem with a closed-form solution. We find that $M=1000$ centers suffice for our robot, in agreement with the analysis in \cite{kaundinya2026precision}.

\subsubsection{Learning the aSSM parametrization and reduced dynamics}
We center each decay about its static configuration to obtain $\{\mathbf{s}^i_{\mathrm{centered}}, \mathbf{u}^i_{\mathrm{decay}}\}_{i=1}^{N}$, then delay embed this data $p$ times through the map $\boldsymbol{\mu}$ presented in Sec.~\ref{subsec:theory_MPC}. An economy SVD of the embedded decays gives chart maps $\{\mathbf{V}^i\}_{i=1}^{N}$, interpolated over $\mathbf{u}_{\mathrm{static}}$ or, when local charts nearly coincide, replaced by a single global chart. The reduced-coordinate data is $\boldsymbol{\mathcal{R}}^i = (\mathbf{V}^i)^{\mathrm T} \boldsymbol{\mu}(\mathbf{s}^{i}_{\mathrm{centered}})$ for the $i^{\text{th}}$ decay; finite differencing gives the velocity data $\dot{\boldsymbol{\mathcal{R}}}^i$.

We seek the leading-order aSSM parametrization $\mathbf{W}(\cdot)$ and reduced dynamics $\mathbf{R}(\cdot)$ as polynomial expansions,
\begin{align}
\label{eq:W_poly_MPC}
\mathbf{W}(\mathbf{r}, \mathbf{s}) &= \sum_{1 \leq |\boldsymbol{\alpha}| \leq k^W_r} \sum_{0 \leq |\boldsymbol{\beta}| \leq k^W_u} \mathbf{W}_{\boldsymbol{\alpha}\boldsymbol{\beta}}\, \mathbf{r}^{\boldsymbol{\alpha}}\,\mathbf{s}^{\boldsymbol{\beta}}, \\
\label{eq:R_poly_MPC}
\mathbf{R}(\mathbf{r}, \mathbf{s}) &= \sum_{1 \leq |\boldsymbol{\alpha}| \leq k^R_r} \sum_{0 \leq |\boldsymbol{\beta}| \leq k^R_u} \mathbf{R}_{\boldsymbol{\alpha}\boldsymbol{\beta}}\, \mathbf{r}^{\boldsymbol{\alpha}}\, \mathbf{s}^{\boldsymbol{\beta}},
\end{align}
with $\boldsymbol{\alpha}$, $\boldsymbol{\beta}$ multi-indices over $\mathbf{r}$ and $\mathbf{s}$, $k^W_r, k^W_u$ and $k^R_r, k^R_u$ the corresponding polynomial orders, and $\mathbf{W}_{\boldsymbol{\alpha}\boldsymbol{\beta}}$, $\mathbf{R}_{\boldsymbol{\alpha}\boldsymbol{\beta}}$ the unknown coefficients. We fit these by minimizing the reconstruction and dynamics regression losses on the centered decay data,
\begin{align}
\label{eq:learn_aSSM_MPC}
\mathbf{W}^{\star} &= \operatorname*{arg\,min}_{\mathbf{W}}\ \sum_{i=1}^{N} \big\| \mathbf{s}^i_{\mathrm{centered}} - \mathbf{W}\big(\boldsymbol{\mathcal{R}}^i, \mathbf{s}^i_{\mathrm{static}}\big) \big\|^2, \\
\mathbf{R}^{\star} &= \operatorname*{arg\,min}_{\mathbf{R}}\ \sum_{i=1}^{N} \big\| \dot{\boldsymbol{\mathcal{R}}}^i - \mathbf{R}\big(\boldsymbol{\mathcal{R}}^i, \mathbf{s}^i_{\mathrm{static}}\big) \big\|^2. \nonumber
\end{align}
The parameters concerning the aSSM-reduced model are chosen to minimize the normalized mean trajectory error on unseen test decay data. We observe that a $d=6$ dimensional aSSM-reduced model with $k^W_r =k^W_u =k^R_r =k^R_u = 2$ provides the lowest errors on the held-out decay data.

\section{Results and Discussion}
\begin{table}[!t]
\renewcommand{\arraystretch}{1.1}
\caption{Open-loop prediction error of the reduced-order models. Autoregressive rollout
from a single initial condition for 750 time steps.
Evaluated on the 10 recorded sequences.}
\label{tab:openloop_error}
\centering
\footnotesize
\setlength{\tabcolsep}{4pt}
\begin{tabular}{lccc}
\toprule
& \multicolumn{3}{c}{Position error (mm)} \\
\cmidrule(lr){2-4}
Model & {Body 1} & {Body 3} & {Body 5 (Tip)}\\
\midrule
aSSM (6D)   & \textbf{1.5}\,$\pm$\,0.8 & \textbf{15.4}\,$\pm$\,7.2 & \textbf{27.5}\,$\pm$\,12.0 \\
ERA (4D)    & 3.4\,$\pm$\,1.4          & 31.1\,$\pm$\,9.0        & 80.6\,$\pm$\,27.4    \\
DMDc (10D)  & 8.6\,$\pm$\,4.2       & 50.2\,$\pm$\,21.5        & 106.4\,$\pm$\,42.3    \\
LOpInf (4D)  & 23.4\,$\pm$\,15.6         & 115.7\,$\pm$\,72.8      & 261.2\,$\pm$\,182.7 \\
\bottomrule
\end{tabular}
\begin{flushleft}
\footnotesize Mean $\pm$ std are taken over all time steps across the
10 sequences. Prediction error is
$\lVert \hat{\mathbf{p}}-\mathbf{p}\rVert$. Parentheses give the reduced model
dimension. Best per column in bold.
\end{flushleft}
\vspace{-1em}
\end{table}

\begin{figure}[!t]
\centering
\includegraphics[width=\columnwidth]{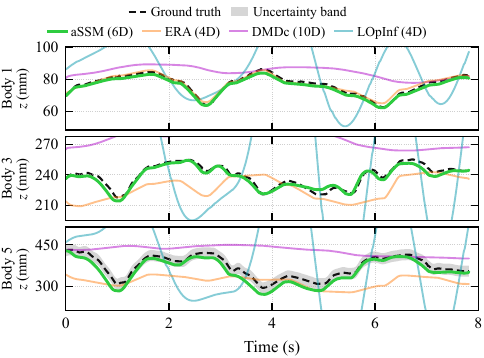}
\caption{Open-loop height $z$ prediction on a sample sequence for bodies 1, 3 and 5 (tip). Each model is rolled out from a single initial condition for 750 time steps with no ground truth measurement feedback. Height $z$ is measured from the base (body~0) along the trunk axis. The shaded band is the $\pm20$~mm measured repeatability of the platform, combining material hysteresis and motor tracking error.}
\label{fig:openloop_height}
\vspace{-1em}
\end{figure}

\begin{figure*}[t]
  \centering
  \includegraphics[width=\textwidth]{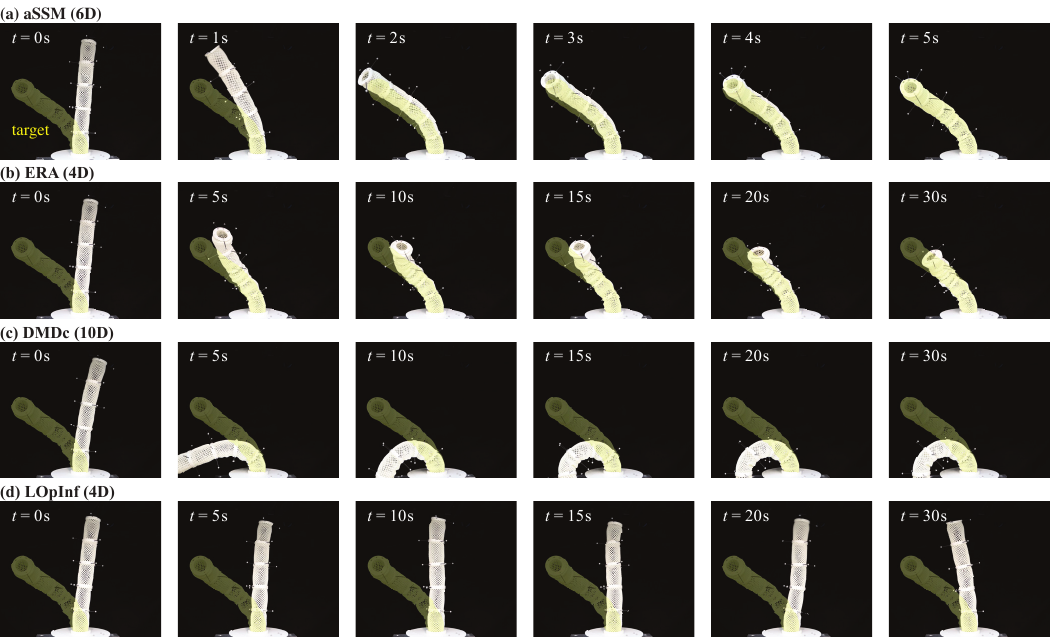}
  \caption{%
    Closed-loop shape control based on \textbf{(a)} aSSM, \textbf{(b)} ERA, \textbf{(c)} DMDc, and \textbf{(d)} LOpInf toward one of the 8 targets. 
    The physical trunk is shown in white and the yellow silhouette shows the target, which is independently recorded by human teleoperation of the trunk. 
    The aSSM is shown over $0$--$5\,\mathrm{s}$ and reaches the target within $4.2\,\mathrm{s}$; the three linear baselines are shown over the full $30\,\mathrm{s}$ hold and none of them converges to the target shape.}
  \label{fig:shape_convergence}
  \vspace{-1em}
\end{figure*}

We evaluate the aSSM-reduced model for shape-control tasks on our hardware soft-trunk platform. We further compare our method against three linear data-driven baselines: the eigensystem realization algorithm (ERA) \cite{juang85}, dynamic mode decomposition with control (DMDc) \cite{Proctor2016}, and the Lagrangian operator inference (LOpInf) \cite{sharma_LpOpInf, sharma_IFAC}. These baselines are chosen following a recent comparative study of shape control \cite{Adibnazari2025}. All four models are identified from the same training data of $1.79\times10^{6}$ time steps at \SI{100}{\hertz}, just under 5 hours of robot data, and predict the full pose $\mathbf{s}$ of Eq.~\eqref{eq:pose}.
For each baseline we sweep the reduced order from 2 to 15 and report the dimension with the lowest open-loop mean position error on a separate validation set of $4.35\times10^{5}$ time steps.
Performance degrades with higher dimensions, so we did not explore further.
Open-loop model prediction is evaluated at \SI{10}{\milli\second}, while inside the MPC horizon all four models propagate at the \SI{50}{\milli\second} control step.

\subsection{Open-loop prediction}
Table~\ref{tab:openloop_error} reports prediction error across 10 held-out test sequences, each a $750$-step rollout with $\Delta t =$\SI{10}{ms} from a single initial condition, with no measurement feedback for the full horizon. 
The aSSM is most accurate on every metric. 
Its mean tip error of $27.5$\,mm is $2.9\times$ smaller than ERA (4D), the strongest of the linear baselines.

Fig.~\ref{fig:openloop_height} shows the open-loop prediction of each model on one recorded sequence, for the height $z$ of bodies~1, 3 and~5, the distal ends of the three actuated sections. 
The axial coordinate $z$ is shown because it is the most demanding axis, depending nonlinearly on the lateral deflection.
The aSSM prediction follows the ground truth closely at all three bodies over the full horizon. 
The best-performing linear baseline ERA reproduces body~1 reasonably well but degrades at bodies~3 and 5. 
Its prediction retains the slow trend while attenuating the oscillations and settling to a systematic offset. 
The DMDc prediction is nearly flat, drifting slowly and reproducing almost none of the dynamics at any body, while the LOpInf prediction oscillates with growing amplitude.

\begin{table}[!t]
\renewcommand{\arraystretch}{1.1}
\caption{Closed-loop shape position error.}
\label{tab:shape}
\centering
\footnotesize
\setlength{\tabcolsep}{3pt}
\begin{tabular}{lcccc}
\toprule
& \multicolumn{4}{c}{Position error (mm)} \\
\cmidrule(lr){2-5}
Model & {Body 1} & {Body 3} & {Body 5 (Tip)} & {All bodies} \\
\midrule
aSSM (6D)   & \textbf{5.4}\,$\pm$\,2.8 & \textbf{6.4}\,$\pm$\,3.7 & \textbf{3.6}\,$\pm$\,2.4 & \textbf{5.1} \\
ERA (4D)    & 9.0\,$\pm$\,5.5          & 50.4\,$\pm$\,32.7        & 108.9\,$\pm$\,75.4       & 56.1         \\
DMDc (10D)  & 20.0\,$\pm$\,11.8        & 48.7\,$\pm$\,28.6        & 150.9\,$\pm$\,91.9       & 73.2         \\
LOpInf (4D)  & 22.1\,$\pm$\,8.3         & 114.9\,$\pm$\,50.4       & 266.4\,$\pm$\,126.3      & 134.4        \\
\bottomrule
\end{tabular}
\begin{flushleft}
\footnotesize Mean $\pm$ std over all time steps in the
final \SI{4}{\second} of each hold, across the 8 target shapes. Best per
column in bold.
\end{flushleft}
\vspace{-1em}
\end{table}

\subsection{Closed-loop shape control}
\label{sec:shapecontrol}
An MPC controller built on each learned model controls the soft trunk's shape, represented by the poses of bodies~1, 3 and 5, the distal ends of the three independently actuated sections.
The controller is sampling-based, scoring three iterations of \num{1000} candidate inputs against the model prediction over a \SI{50}{\milli\second} horizon and resampling around the best with a shrinking radius.
We evaluate the closed-loop performance on two tasks.
In \emph{smooth and rapid shape regulation}, the target shape is defined by the fixed positions of the three bodies.
In \emph{dynamic shape tracking}, the trunk follows a continuously deforming shape trajectory, tracking both the position and orientation of the three bodies.
The controllers based on all four models drive the same plant at \SI{20}{\hertz} from the same relaxed initial configuration. For brevity we refer to the MPC controller built on each model by the model's name.

\begin{figure*}[t]
  \centering
  \includegraphics[width=\textwidth]{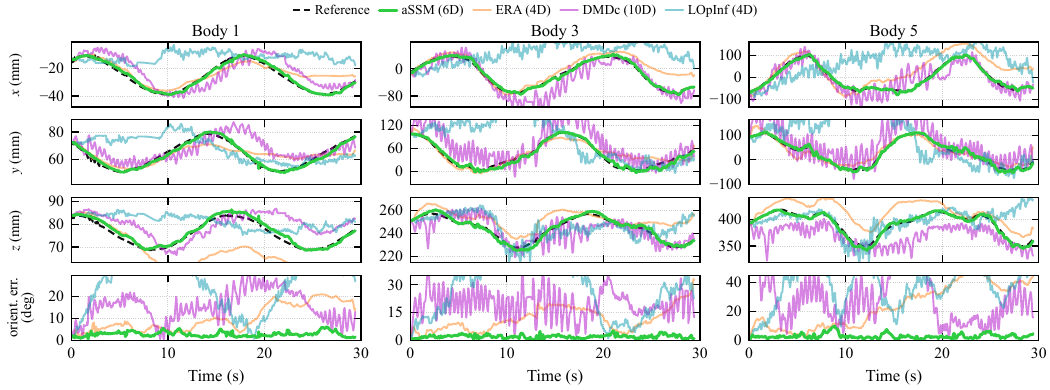}
  \caption{%
    Closed-loop dynamic shape tracking of a \SI{30}{\second} reference trajectory.
    \textbf{Column}: tracked body~1, 3, and~5 (tip) along the trunk. 
    \textbf{Row}: component of that body's pose.
    The top three rows give position along the $x$, $y$ and $z$ axes, with the dashed black line the reference and the colored lines the four models; the bottom row gives the orientation tracking error, the geodesic angle between the measured and the reference orientation.}
  \label{fig:shape_tracking}
  \vspace{-1em}
\end{figure*}

\subsubsection{Smooth and rapid shape regulation}
\label{sec:shapestatic}
We evaluate the aSSM and baseline models on the same set of 8 target shapes.
The target shapes are obtained by teleoperating the soft trunk to diverse configurations and recording the resulting body poses, which ensures target reachability.
A trial is declared successful once the position error $\lVert \mathbf{p}_{\mathrm{target},i}-\mathbf{p}_i\rVert$, averaged over bodies~1, 3, and 5, falls below \SI{10}{\milli\metre} and is held for \SI{4}{\second}; an unsuccessful trial is terminated after \SI{30}{\second}. 
The aSSM succeeds on all 8 shapes, reaching the target at \SI{11.8}{\second} on average (between \SI{4.2}{\second} and \SI{25.6}{\second}).
In contrast, none of the three linear baselines reaches the target shape. This can be reasoned from the nonlinear behavior arising from vertical deformations of the soft trunk, already observed in \cite{Kaundinya2026} and \cite{kaundinya2026precision} for similar trunk geometries at the tip positions. Our analysis thus demonstrates that a nonlinear, low-dimensional aSSM model accurately tracks these shapes in real time.

Table~\ref{tab:shape} reports the position error of each body, averaged over the last \SI{4}{\second} of each trial.
The aSSM-based MPC controller outperforms the baselines on every body across the 8 target shapes. 
It reaches a mean per-body error of $5.1$\,mm, $10.9\times$ smaller than ERA ($56.1$\,mm), the strongest of the linear baselines, $14.2\times$ smaller than DMDc ($73.2$\,mm) and $26.2\times$ smaller than LOpInf ($134.4$\,mm). 
The error distribution along the trunk shows that the aSSM model tracks the full body shape with comparable accuracy at every body, whereas the linear models' error grows monotonically from base to tip.
This reflects the limitation of a single linear operator fit across the whole workspace, which cannot capture a map whose nonlinearity grows with large deflection.

\begin{table}[!t]
\renewcommand{\arraystretch}{1.1}
\caption{Closed-loop dynamic shape tracking error.}
\label{tab:shapetrack}
\centering
\footnotesize
\setlength{\tabcolsep}{2pt}
\begin{tabular}{lcccc}
\toprule
& \multicolumn{3}{c}{Position error (mm)} & {Orientation error (\si{\degree})} \\
\cmidrule(lr){2-4} \cmidrule(lr){5-5}
Model & {Body 1} & {Body 3} & {Body 5 (Tip)} & {All bodies} \\
\midrule
aSSM (6D)   & \textbf{2.4}\,$\pm$\,0.8 & \textbf{6.4}\,$\pm$\,3.5 & \textbf{7.9}\,$\pm$\,4.3 & \textbf{2.6}\,$\pm$\,1.3 \\
ERA (4D)    & 11.9\,$\pm$\,5.8         & 24.5\,$\pm$\,16.1        & 56.1\,$\pm$\,21.9        & 15.1\,$\pm$\,11.6       \\
DMDc (10D)  & 12.3\,$\pm$\,5.4         & 38.5\,$\pm$\,20.7        & 61.9\,$\pm$\,33.3        & 21.7\,$\pm$\,13.5         \\
LOpInf (4D)  & 20.3\,$\pm$\,9.6         & 97.1\,$\pm$\,49.6        & 150.7\,$\pm$\,95.1       & 32.2\,$\pm$\,18.2       \\
\bottomrule
\end{tabular}
\begin{flushleft}
\footnotesize Mean $\pm$ std over all time steps. Best per column in bold.
\end{flushleft}
\vspace{-1em}
\end{table}

Fig.~\ref{fig:shape_convergence} shows convergence to one target in the set. 
The aSSM reaches the target in \SI{4.2}{\second} with a steady-state error of \SI{2.1}{\milli\metre}. 
ERA achieves low position error of \SI{0.5}{\milli\metre} at body~1 but cannot generate the distal curvature, with \SI{149.8}{\milli\metre} error at the tip.
DMDc drives the trunk into a spurious arch, while LOpInf commands the least
actuation and leaves the trunk near its undeflected configuration.

\subsubsection{Dynamic shape tracking}
\label{sec:shapedynamic}
To further test the accuracy of the modeled dynamics, we record a continuously varying shape trajectory by driving each section with a phase-shifted sinusoid over a \SI{15.8}{\second} period.
The resulting \SI{30}{\second} reference trajectory has every section both bending and rotating, sweeping bodies~1, 3 and~5 by $29$, $109$ and $167$\,mm peak to peak.

Table~\ref{tab:shapetrack} reports the position and orientation tracking error. 
The aSSM follows the moving shape to a mean per-body error of $5.6$\,mm, comparable to the $5.1$\,mm it achieves on fixed target shapes and far better than $30.8$\,mm for ERA, $37.6$\,mm for DMDc, and $89.4$\,mm for LOpInf.
Orientation error is flat along the trunk at 2.8, 2.0 and 2.9$^\circ$ at bodies 1, 3 and 5.
For ERA, it grows from 10.1$^\circ$ at body 1 to 23.8$^\circ$ at the tip.
Fig.~\ref{fig:shape_tracking} shows the nine tracked position axes: the aSSM follows the reference trajectory, ERA reproduces the shape of the motion with an amplitude and offset error that grows distally, and DMDc and LOpInf oscillate heavily.

\begin{figure*}[t]
\centering
\includegraphics[width=\textwidth]{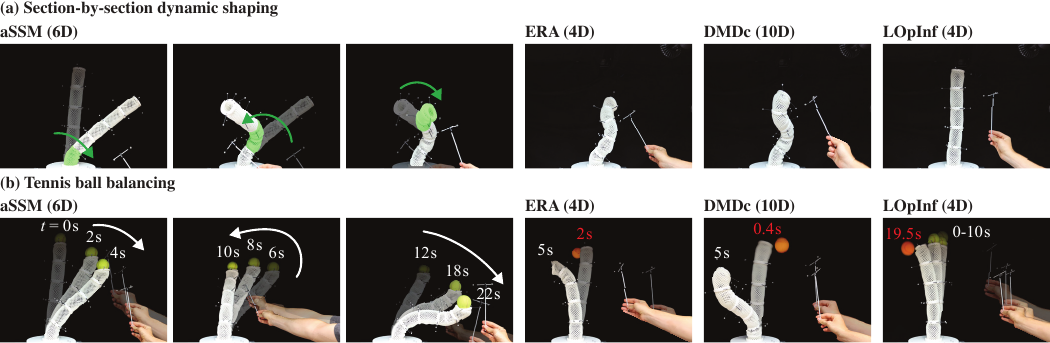}
\caption{Human-in-the-loop shape control tasks under each learned model, all driven by the hand-held wand.
\textbf{(a)} Section-by-section dynamic shaping: the operator selects one section at a time, and the active section (green) holds its distal body at a \SI{10}{\centi\metre} standoff from the wand while the remaining sections hold position. 
Three frames show the aSSM driving each section; the linear baselines are shown at their final reached shape.
\textbf{(b)} Tennis ball balancing: the tip follows the wand with the same \SI{10}{\centi\metre} standoff while its orientation is constrained to keep the distal face level.
Labels mark the time along the run, while red marks the instant the ball leaves the face. 
Under the aSSM the ball is held for over \SI{25}{\second}.
ERA and DMDc hold neither the standoff nor the face level and lose the ball after \SI{2}{\second} and \SI{0.4}{\second}; LOpInf keeps it to \SI{19.5}{\second} only by leaving the trunk essentially undeflected.}
\label{fig:demos}
\vspace{-1em}
\end{figure*}

\subsection{Dynamic human interaction}
\label{sec:demos}
Going beyond pre-recorded target shapes and reference trajectories, we now place a human in the loop to provide the reference with a motion-captured wand.
Three tasks are considered: the trunk must follow the wand at the tip, follow it one section at a time, and follow it while additionally holding the tip orientation level, so that an object can be balanced on the distal face.
The controllers based on all four models drive the same plant
at 20 Hz.

\subsubsection{Tip position tracking}
The human operator waves the wand freely in space, so the resulting motion is not confined to the trunk's reachable workspace.
We record 10 trajectories of \SI{20}{\second} each and replay them as the reference trajectory for tip position tracking, with all four models evaluated on identical references.
The aSSM tracks the wand to a mean tip error of \SI{36.2}{\milli\metre}, $1.6\times$ smaller than ERA (\SI{59.3}{\milli\metre}), $3.0\times$ smaller than DMDc (\SI{106.9}{\milli\metre}) and $6.8\times$ smaller than LOpInf (\SI{247.0}{\milli\metre}).
The error is above the \SI{7.9}{\milli\metre} the aSSM reaches at the tip in Sec.~\ref{sec:shapecontrol} and reflects the unreachable portions of the reference, which no model can track.
The qualitative behavior matches Sec.~\ref{sec:shapecontrol}: ERA follows the motion at a distally growing offset, DMDc oscillates about the reference, and LOpInf leaves the trunk near undeflected.

\subsubsection{Section-by-section dynamic shaping} 
This task exercises the actuation decoupling shown in Sec.~\ref{sec:platform}.
With our tendon routing design, a single section can now be driven while the others hold their previous commands.
The human operator shapes one section at a time (Fig.~\ref{fig:demos}a), selecting the active section and thereby gating which group of tendon motors is driven.
The active section holds its distal body at a \SI{10}{\centi\metre} standoff from the wand, which makes both attracting and repelling the trunk possible.

Fig.~\ref{fig:demos}a shows aSSM driving sections 1 to 3 in turn, with the active section highlighted in green.
The decoupling is directly visible: the inactive sections hold their shape while the active one is reshaped.
The three baselines are shown at the final shape each reaches.
ERA and DMDc respond to the wand and reach shapes commanded by the operator, whereas LOpInf barely reacts and leaves the trunk near undeflected.

\subsubsection{Balancing a tennis ball} 
A tennis ball is placed on the distal ring of the trunk.
The tip follows the wand at the same \SI{10}{\centi\metre} standoff while its orientation is additionally constrained to keep that ring level, so the ball stays balanced.

Fig.~\ref{fig:demos}b compares the performance of the four models.
With the aSSM, the tennis ball stays balanced throughout the interaction, for over \SI{25}{\second}.
ERA and DMDc hold neither the standoff nor the level, tilting the tip as the trunk bends, and lose the ball after \SI{2}{\second} and \SI{0.4}{\second}, respectively.
LOpInf keeps it to \SI{19.5}{\second}, but again only by not reacting to the wand leaving the trunk essentially undeflected.

The controllers considered so far all rely on a learned model.
To test whether the same task can be solved without one, we additionally train a model-free policy directly on the hardware with Proximal Policy Optimization (PPO)~\cite{schulman2017proximal}.
The policy observes the wand-to-tip geometry and the tip normal over a short history, together with the current motor positions and the previous action.
It commands normalized motor velocities for the nine tendons at \SI{100}{\hertz}, five times the MPC rate, projected onto the admissible set defined by the limits in Sec.~\ref{sec:platform}.
The main reward terms are proximity to the stand-off and a level tip, with penalties on actuation.
Training runs for 4320 iterations, corresponding to $4.3\times10^{6}$ environment steps and \SI{12}{\hour} of robot interaction.

\begin{figure}[t]
  \centering
  \includegraphics[width=\columnwidth]{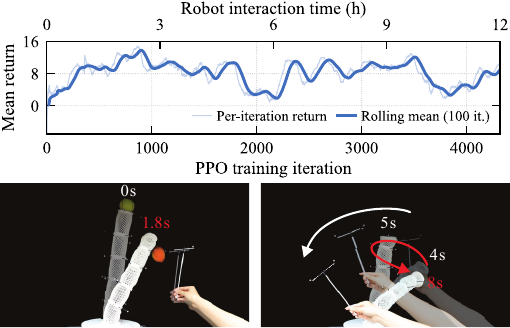}
  \caption{Model-free reinforcement learning on hardware for the tennis ball balancing task. 
  \textbf{Top}: mean return over 4,320 PPO iterations trained directly on the soft trunk, corresponding to \SI{12}{\hour} of robot interaction. 
  \textbf{Bottom}: the resulting policy evaluated over \SI{8}{\second}.
  Left: the trunk drops the ball after \SI{1.8}{\second}. 
  Right: the tip tracks the wand only over a small region of the workspace (\SIrange{4}{5}{\second}) and moves to the opposite side as the wand travels beyond it.
  The white arrow shows the wand motion and the red arrow the tip motion.}
  \label{fig:rl}
  \vspace{-1em}
\end{figure}

Fig.~\ref{fig:rl} shows the mean return rising over the first 900 iterations and then oscillating without further improvement, and the behavior of the resulting policy.
The task requires the tip to both follow the wand and hold its distal face level, and the policy achieves neither.
The trunk drops the ball after \SI{1.8}{\second}, and the policy tracks the wand only over a small region of the workspace, driving the trunk to the opposite side once the wand leaves that region.
We report this as a sample-efficiency result rather than a comparison of methods: under a hardware data budget of \SI{12}{\hour} the model-free policy does not reach a working solution on this task, whereas the aSSM-reduced model is identified from \SI{5}{\hour} of decay data and supports every controller in this paper, including the one that balances the ball for over \SI{25}{\second}.

\section{Conclusions}

In this paper, we have demonstrated real-time closed-loop shape control of a long helicoid trunk robot via an aSSM-based MPC scheme. Our methodology extends prior work on aSSM modeling to accurately model the positions and orientations of bodies along the trunk. Specifically, we learned a 6D aSSM-reduced model for the full trunk shape profile, which predicts on previously unseen data with consistent accuracy and outperforms available linear, data-driven methods for modeling soft robotic shape profiles. We then ran closed-loop experiments using our aSSM-reduced models within standard MPC schemes to track the dynamic shape of the trunk, follow a user's hand as they navigate the workspace, reshape one section at a time while the others hold, and balance a tennis ball on the tip while it follows the hand. Across these tasks, our aSSM-based MPC controller outperforms MPC controllers built on linear data-driven models, and solves the ball-balancing task that a model-free RL policy fails to solve with more than twice the hardware interaction time.

Our approach is based on dynamic assumptions that are routinely observed to hold for soft robots. Hence, it offers a general methodology applicable to other soft robotic systems with finite-dimensional actuation spaces. Our helicoid trunk robot was shown to deform into interesting shapes and dynamically track them, but the dynamic shape-changing rate is still limited by tendon speed. Future work will explore upgrading the actuation system to enable faster dynamic shape tracking. Beyond the demonstrated tasks, we hope to leverage our accurate shape tracking via aSSM-based MPC controllers to enable practical applications such as navigating and exploring confined environments where a soft manipulator's full-body pose matters.






\bibliographystyle{./IEEEtranBST/IEEEtran}
\bibliography{./root,aperiodic_thesis_bib_2}

\end{document}